\documentclass[lettersize,journal]{IEEEtran}
\usepackage{amsmath,amsfonts}
\usepackage{algorithmic}
\usepackage{algorithm}
\usepackage{array}
\usepackage[caption=false,font=normalsize,labelfont=sf,textfont=sf]{subfig}
\usepackage{textcomp}
\usepackage{stfloats}
\usepackage{url}
\usepackage{verbatim}
\usepackage{graphicx}
\usepackage{cite}
\usepackage{threeparttable}

\begin{document}
\title{Alleviating neighbor bias: augmenting graph self-supervise learning with structural equivalent positive samples}

\author{Jiawei Zhu, Mei Hong, Ronghua Du, Haifeng Li
\IEEEcompsocitemizethanks{\IEEEcompsocthanksitem J. Zhu and H. Li are with the School of Geosciences and Info-Physics, Central South University, Changsha 410083, China (e-mail: lihaifeng@csu.edu.cn)\
\IEEEcompsocthanksitem M. Hong is with Institute of Meteorology and Oceanography, National University of Defence Technology, Changsha 410073, China\
\IEEEcompsocthanksitem R. Du is with the College of Automotive and Mechanical Engineering, Changsha University of Science and Technology, Changsha 410114, China.
}
\thanks{\textit{Corresponding author: Haifeng li.}}
}

\markboth{Journal of \LaTeX\ Class Files,~Vol.~14, No.~8, August~2015}%
{Shell \MakeLowercase{\textit{et al.}}: Bare Advanced Demo of IEEEtran.cls for IEEE Computer Society Journals}

\IEEEtitleabstractindextext{%
\begin{abstract}
In recent years, using a self-supervised learning framework to learn the general characteristics of graphs has been considered a promising paradigm for graph representation learning. The core of self-supervised learning strategies for graph neural networks lies in constructing suitable positive sample selection strategies. However, existing GNNs typically aggregate information from neighboring nodes to update node representations, leading to an over-reliance on neighboring positive samples, i.e., homophilous samples; while ignoring long-range positive samples, i.e., positive samples that are far apart on the graph but structurally equivalent samples, a problem we call "neighbor bias." This neighbor bias can reduce the generalization performance of GNNs. In this paper, we argue that the generalization properties of GNNs should be determined by combining homogeneous samples and structurally equivalent samples, which we call the "GC combination hypothesis." Therefore, we propose a topological signal-driven self-supervised method. It uses a topological information-guided structural equivalence sampling strategy. First, we extract multiscale topological features using persistent homology. Then we compute the structural equivalence of node pairs based on their topological features. In particular, we design a topological loss function to pull in non-neighboring node pairs with high structural equivalence in the representation space to alleviate neighbor bias. Finally, we use the joint training mechanism to adjust the effect of structural equivalence on the model to fit datasets with different characteristics. We conducted experiments on the node classification task across seven graph datasets. The results show that the model performance can be effectively improved using a strategy of topological signal enhancement. 

\end{abstract}
\begin{IEEEkeywords}
Graph self-supervised Learning, Structural equivalence, Graph convolutional network, Persistent homology.
\end{IEEEkeywords}}

\maketitle

\IEEEdisplaynontitleabstractindextext

%
\IEEEpeerreviewmaketitle

\ifCLASSOPTIONcompsoc
\IEEEraisesectionheading{\section{Introduction}\label{sec:introduction}}
\else
\section{Introduction}
\label{sec:introduction}
\fi
\IEEEPARstart{I}{n} recent years, self-supervised learning (SSL) has been introduced into the image\cite{kolesnikov2019revisiting,doersch2015unsupervised,caron2018deep,misra2020self} and text\cite{le2014distributed,devlin2018bert,lewis2019bart,lan2019albert} domains. SSL mines supervised information from large-scale unsupervised data and trains the network with this constructed supervised information. It has now become a standard representation learning paradigm, reaching or even exceeding the performance of pre-trained models via supervised learning\cite{tian2020contrastive,he2020momentum,chen2020simple,grill2020bootstrap}. Many works attempt to migrate SSL to graph data to facilitate learning tasks on graphs, such as node classification and graph categorization\cite{wu2020comprehensive}. SSL can also mitigate the label dependence, poor generalization, and weak robustness problems associated with semi-supervised/supervised learning-style graph neural networks (GNNs).

However, there are several critical challenges in applying SSL on GNNs due to the uniqueness of graph-structured data. First, graph-structured data are often more complex than other domains (e.g., images and text)\cite{jin2020self}. In addition to node attributes, graphs possess complex structural information. For example, while the topology of an image is a fixed grid and text is a simple sequence, graphs are not limited to these rigid structures. Moreover, the entire structure is a single data instance for image and text, while each node in the graph is a separate instance with its associated properties and topology. The complexity of graph-structured data does not end there. Text and image domains normally assume data samples are independent and identically distributed. However, instances (or nodes) are inherently interconnected and interdependent in the graph domain. The complex nature of graph-structured data makes it challenging to apply self-supervised learning developed in other domains to graphs directly. But this complexity also provides a wealth of information that allows us to design pretext tasks from different perspectives.

Therefore, to obtain a good representation of the graph and perform effective pre-training, self-supervised models should obtain basic information from the graph's node properties and structural topology\cite{ma2021deep}. One of the main directions is to construct self-supervised information for nodes based on their local structural information, or their relationships with the rest of the graph. Currently, most graph contrastive learning, in order to take into account the topology, generally takes the subgraphs derived from nodes and their neighborhood as instances, and then different methods use different strategies to perturb the instances on nodes, edges and features to generate different views, and then maximize the consistency between different views to learn the representations. In other words, by introducing different perturbations to different data augmentation schemes to generate samples of different views, the model is able to learn invariant feature representation.

Even though the contrastive mechanisms of graph SSL now vary, most adopt a message passing framework, in which aggregating neighboring features (messages) and updating own node features are the two core procedures. This framework assumes that the data is homophilous (i.e., the features or labels of neighboring nodes are similar or smoothed) so that satisfied performance on downstream tasks can be obtained by using neighboring nodes to update specific node features\cite{wang2021powerful}. However, our statistics on commonly used realistic datasets reveal that the homophily index varies across datasets, as shown in Table \ref{tbl:1}. The homophily mechanism cannot explain all positive samples in the data; a certain number of distant positive sample pairs still exist in the dataset; we name this phenomenon neighbor bias.
\begin{table}[h]
  \centering
  \caption{Homophily index of datasets}
  \label{tbl:1}
  \begin{tabular}{cc} 
  \hline
  Dataset          & Homophily index  \\ 
  \hline
  Cora             & 0.8138                                                    \\
  Citeseer         & 0.7461                                                    \\
  PubMed           & 0.8024                                                    \\
  Amazon Photo     & 0.8306                                                    \\
  Amazon Computers & 0.7864                                                    \\
  Coauthor Physics & 0.9314                                                    \\
  Coauthor CS      & 0.8081                                                    \\
  \hline
  \end{tabular}
  \end{table}

Therefore, this study argues that there are other mechanisms for deriving positive samples on graphs besides the homophily assumption, and structural equivalence is one of them. That is, the label consistency of certain nodes in these datasets is defined by their structural equivalence. For example, we expect social networks to exhibit strong ties based on similar attributes; however, there may also be "familiar strangers," central nodes that do not interact but share common interests, which both appear as star-like centroids in the network and exhibit structural equivalence.

Although it is possible to enable the learning process to use distant positive samples by deepening the graph neural network model, this approach has two problems. One is that too deep graph convolutional neural networks can lead to over-smoothness or bottleneck. The other is that this does not differentiate information among remote nodes, which may lead to misuse in distant negative samples. While other existing studies model structural similarity, most use simple topological features (e.g., node degree, etc.), which cannot fully capture the local structure of nodes, and many common topological descriptions fail in the case. For example, nodes have the same degree may still different local structure, the degree of the central node cannot tell whether its neighbor nodes are connected with each other or not. Therefore, we introduce persistent homology to describe the local structure of the nodes, which proved can be used to study complex topologies at multiple scales.

We perform statistics on the positive samples that do not satisfy the homophily assumption, and the results are shown in Table \ref{tbl:2}. It can be seen that non-neighbor samples on most datasets are far apart on the graph. When we compare the structural equivalence of node pairs using the method based on persistent homology, we find that basically, the topological distance between non-neighbor negative samples is larger than the average topological distance; and the topological distance between non-neighbor positive samples is smaller than the average topological distance. Meanwhile, there is a trend that the larger the hop distance, the smaller the topological distance. This verifies our hypothesis and shows that the positive samples ignored under the homophily assumption can be derived based on the hop count and structural equivalence restrictions.
\begin{table*}
  \centering
  \caption{Topological distance between non-neighboring samples}
  \label{tbl:2}
  \begin{tabular}{cccccccc} 
  \hline
  Dataset                                                   & \begin{tabular}[c]{@{}c@{}}Average\\hops\end{tabular} & \begin{tabular}[c]{@{}c@{}}Average\\topological\\distance\end{tabular} & \begin{tabular}[c]{@{}c@{}}Topological distance\\between \\positive samples\\(3 hop)\end{tabular} & \begin{tabular}[c]{@{}c@{}}Topological distance\\between\\positive samples\\(4 hop)\end{tabular} & \begin{tabular}[c]{@{}c@{}}Topological distance\\between\\positive samples\\(5 hop)\end{tabular} & \begin{tabular}[c]{@{}c@{}}Topological distance\\between\\positive samples\\(5 hop)\end{tabular} & \begin{tabular}[c]{@{}c@{}}Topological distance\\between\\negative samples\\(non-neighbor)\end{tabular}  \\ 
  \hline
  Cora                                                      & 5.5                                                   & 0.0456                                                                 & 0.0391                                                                                            & 0.0371                                                                                           & 0.0356                                                                                           & 0.0348                                                                                           & 0.0462                                                                                                   \\
  Citeseer                                                  & 7.0                                                   & 0.0583                                                                 & 0.0588                                                                                            & 0.0546                                                                                           & 0.0515                                                                                           & 0.0483                                                                                           & 0.0640                                                                                                   \\
  PubMed                                                    & 6.0                                                   & 0.0278                                                                 & 0.0264                                                                                            & 0.0243                                                                                           & 0.0200                                                                                           & 0.0171                                                                                           & 0.0278                                                                                                   \\
  \begin{tabular}[c]{@{}c@{}}Amazon\\Computers\end{tabular} & 2.8                                                   & 0.0376                                                                 & 0.0251                                                                                            & 0.0227                                                                                           & 0.0088                                                                                           & 0.0027                                                                                           & 0.0365                                                                                                   \\
  \begin{tabular}[c]{@{}c@{}}Amazon\\Photo\end{tabular}     & 3.2                                                   & 0.3330                                                                 & 0.2346                                                                                            & 0.2408                                                                                           & 0.2838                                                                                           & 0.2367                                                                                           & 0.3364                                                                                                   \\
  \begin{tabular}[c]{@{}c@{}}Coauthor\\Physics\end{tabular} & 4.5                                                   & 0.0945                                                                 & 0.0912                                                                                            & 0.0667                                                                                           & 0.0498                                                                                           & 0.0439                                                                                           & 0.0911                                                                                                   \\
  \begin{tabular}[c]{@{}c@{}}Coauthor\\CS\end{tabular}      & 4.5                                                   & 0.0784                                                                 & 0.0634                                                                                            & 0.0527                                                                                           & 0.0413                                                                                           & 0.0336                                                                                           & 0.0785                                                                                                   \\
  \hline
  \end{tabular}
  \end{table*}

This study proposes a method that captures structural equivalence through persistence homology to extend positive samples and aid graph representation learning. The main contributions of this study are as follows.

\begin{enumerate}
\item We discover the existence of neighbor bias in realistic datasets. Moreover, we observe that the structural equivalence can capture some distant positive samples. That is, though some node pairs with the same label are far apart on the graph, their structural equivalence is higher than the structural equivalence of pairs of nodes with different labels.

\item To alleviate the neighbor bias, this study proposes a topological signal-enhanced graph self-supervised method that captures structural equivalence through topological invariants and draws high structural equivalence node pairs in the feature space, and finally uses an adaptive way to balance structural equivalence and homogeneity to assist graph representation learning.

\item Our method outperforms the baseline approach on seven node classification benchmark datasets, demonstrating that the introduction of structural equivalence can enhance the expressiveness of graph representation learning and showing the great potential that the complex topological signal of the graph itself can bring to graph representation learning.
   \end{enumerate}

The remainder of this paper is organized as follows: in Section 2, we briefly review work related to graph self-supervised learning and existing graph representation learning methods based on structural equivalence. Section 3 briefly introduces the persistent homology and persistence image. Section 4 describes the approach for extracting topologically intrinsic signals, the strategy to construct the positive samples, the loss function of our proposed method, and the training approach. The performance of this method is evaluated experimentally in Section 5. The impact of different topological signal extraction strategies on the model and the importance of topological information on different datasets are discussed in Section 6. The full paper is summarized in section 7.

\section{Related Work}
The great success of self-supervised learning in computer vision and natural language processing has sparked interest in applying it to graph data. However, transferring pretext tasks designed for CV/NLP to the domain of graph learning is not easy. The main challenge is that graphs are irregular non-Euclidean data. Compared to the regular lattice Euclidean space where image/language data are located, non-Euclidean spaces are more general but also more complex. In addition, data instances (nodes) in graph data are naturally related to the topology, whereas instances in CV (images) and NLP (text) tend to be independent. Therefore. How to handle this dependency in graph natural supervised learning becomes a challenge for the design of the pretex task.

Graph Contrastive Learning: in general, self-supervised learning learns representations by implementing proxy goals between inputs and customized signals, where contrastive approaches\cite{tian2020contrastive,he2020momentum,oord2018representation} have achieved impressive performance in learning image representations by maximizing two views (or augmentatations) of the same input. Inspired by the contrast in vision learning success, similar approaches have been applied to learning graph neural networks\cite{hassani2020contrastive,qiu2020gcc}. DGI\cite{velickovic2019deep} relied on a parameterized mutual information estimate to distinguish positive node-graph pairs from negative node-graph pairs; MVGRL\cite{hassani2020contrastive} created views for graphs by introducing graph diffusion and generalizing CMC\cite{tian2020contrastive} to graph-structured data; GCC\cite{qiu2020gcc} uses InfoNCE loss\cite{tian2020contrastive} and MoCo-based negative pooling\cite{he2020momentum} for large-scale GNN pre-training. GRACE\cite{zhu2020deep}, GCA\cite{zhu2021graph} and GraphCL\cite{you2020graph} follow the idea of SimCLR\cite{chen2020simple} and learn node/graph representation by directly using other nodes/graphs as negative samples. Although these models have achieved impressive performance, they require complex designs and architectures. The choice of data augmentation for a given task relies heavily on experimentation. Also, the hyperparameters of each data augmentation exponentially expand the hyperparameter search space. Therefore, we try to generate positive samples based on the topological signals inherent to the graph and reasonably combine the effects of structural equivalence and homogeneity mechanisms, thus avoiding empirical data augmentation.

Structural equivalence: structural equivalence does not assume that vertices are connected, and the basic assumption of structural equivalence is that vertices with similar local structures should be considered similar \cite{qiu2020gcc}. Indicators such as node degree, structural holes \cite{burt2018structural}, k-core \cite{alvarez2005large}, and motif \cite{milo2002network}, can be used to model structural equivalence. However, these features can only describe simple topologies and are not sufficient to describe the full picture of the local topological structure. Therefore, this study uses a persistent homology-based approach to extract multiscale topological features to quantify structural similarity.

\section{Preliminary}
One important tool in the field of computational topology, or topological data analysis, is persistent homology \cite{zomorodian2004computing,edelsbrunner2008persistent}. The first step before applying persistent homology is constructing simplicial complexes for high-dimensional data. Some of the commonly used simplicial complexes are Vietories Rips complex \cite{zomorodian2010fast}, Alpha complex \cite{edelsbrunner1994three}, etc. A simplicial complex is built based on a proximity threshold. Instead of constructing a single complex to approximate the shape of data, persistent homology constructs a series of complexes to explore the data from multiple levels. Taking the Vietoris-Rips complex as an example, the threshold is continuously increased to connect the points of high-dimensional data with a distance less than this threshold to form a simplicial complex under the current threshold. As the threshold grows, a sequence of nested simplicial complex forms. This process is called filtration. After constructing simplicial complexes, homology can be applied to each complex to abstract corresponding topological information. For example, there are different numbers of connected components, rings, and voids in the topological space X, where the simplicial complexes are constructed. These connected components, rings, and hole structures correspond to the homology groups in the topological space. More precisely, the k-dimensional holes in the space generate the homology group $H_k(X)$, where the rank of the group is the kth order Betti number $k$, i.e., the number of $k$-dimensional holes in X.

For each homology rank $k$, the information of the persistent homology can be expressed as a persistence diagram. Persistence diagram is a multiset on a 2D plane. In the filtration process, the birth time $x$ and death time $y$ of each topological invariant, such as connected components, rings, and voids, can be recorded, and a two-dimensional plane multi-set can be generated with $x$ and $y$ as horizontal and vertical coordinates. This multi-set becomes a persistence diagram. Notice that the vanishing time of the topological invariants must be greater than the appearance time so that all points in the PD lie to the upper left of the diagonal function $y = x$. The points in the PD that are far from the diagonal are usually considered significant features, and the points near the diagonal are considered noise. Note that the $H_0$ feature represents a connected component, and there is an $H_0$ feature point that appears at time 0 and continues to positive infinity.

The persistence diagram has an important property, namely stability\cite{cohen2007stability}. This is the theoretical foundation for the wide application of the persistence diagram. Stability means that there is an upper bound on the distance between the persistence diagrams obtained before and after perturbing the original point cloud. That is, the persistence diagram has the ability to resist noise residing in data. However, the persistence diagram as a multi-set is difficult to combine with more widely used machine learning models, so researchers have proposed some transform representations of PD as vector representations, thus expanding their applications. One of the most general transformations is the persistence image. Persistence image is a transformation algorithm proposed by Adams et al.\cite{adams2017persistence}. For the first time, it is proposed to transform a multi-set PD into a vectorized format. Thus, the difficult-to-apply data format is transformed into a generic vector for various downstream algorithms. Experimental results indicate that the transformation of the persistence image has high computational efficiency and also theoretically guarantees the stability of the transformation.

\section{Self-supervised Learning based on topological invariants}
\begin{figure}[htpb]
	\centering\includegraphics[width=0.8\linewidth]{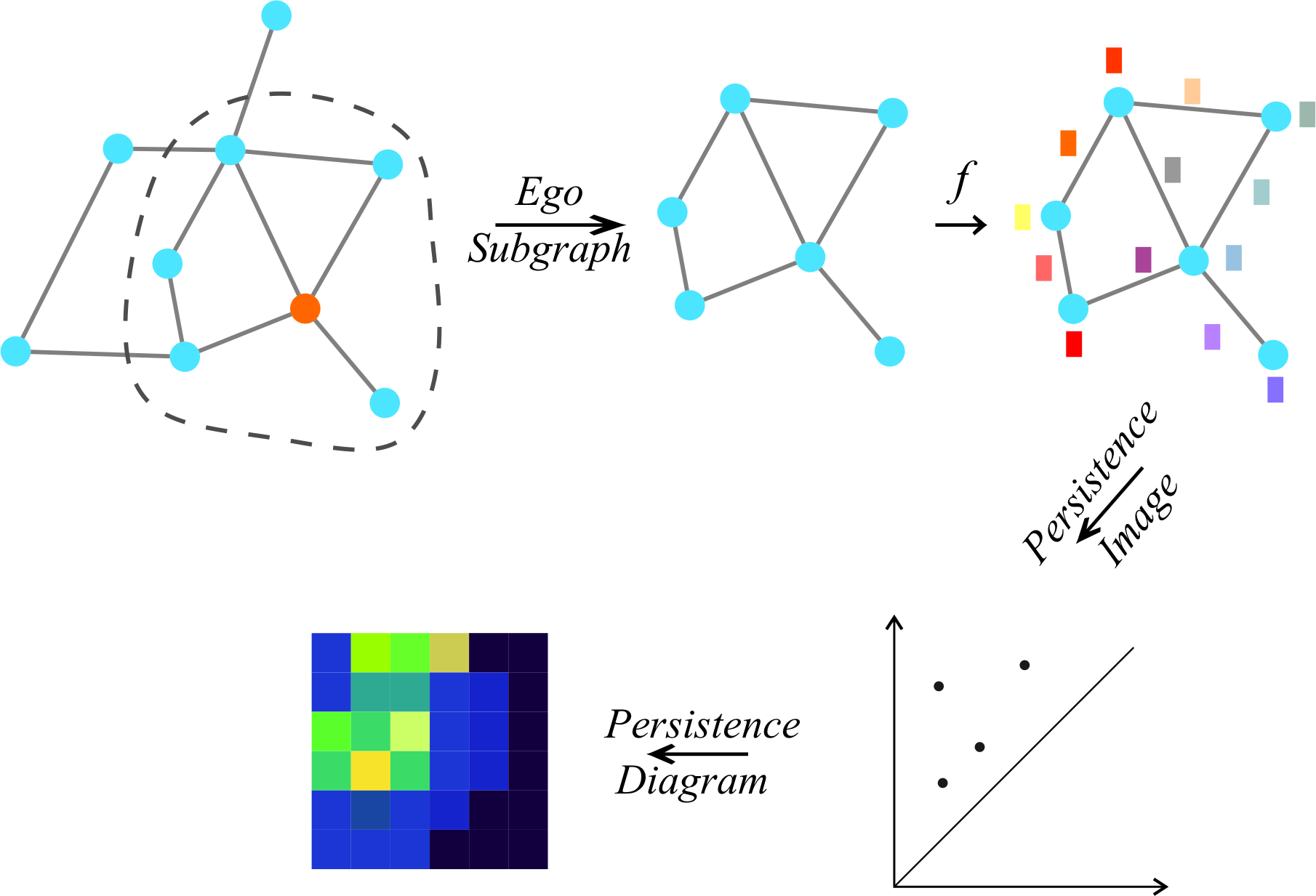}
	\caption{Topological Signal Extraction}
	\label{fig:pi_flow}
\end{figure}

The proposed method has two critical steps: 1) extracting the node's local topological information and 2) implementing the topological signal-based graph contrastive learning. Figure \ref{fig:pi_flow} shows the node's local topological information extraction process. Firstly, the local structure of nodes needs to be extracted. Then, mapping function is defined to assign values to nodes and edges for persistent homology analysis. Finally, the derived persistence diagram is converted into the persistence image as a topological signal readable by the graph neural network.

\subsection{Sub-graph Instance Extraction}
In order to calculate the structural equivalence of nodes, the local topology of the nodes needs to be extracted first. Here we use the ego networks as the local topology of nodes, which consists of a unique central node (ego), and the neighbors (alters) of this node, and the edges include only those between ego and alters and between alters and alters. For a more complete portrayal of the local topological details of the nodes, we will extract the 2-hop ego subgraph (i.e., extending alters to 2-hop neighbors).
\begin{figure}[htpb]
	\centering\includegraphics[width=0.95\linewidth]{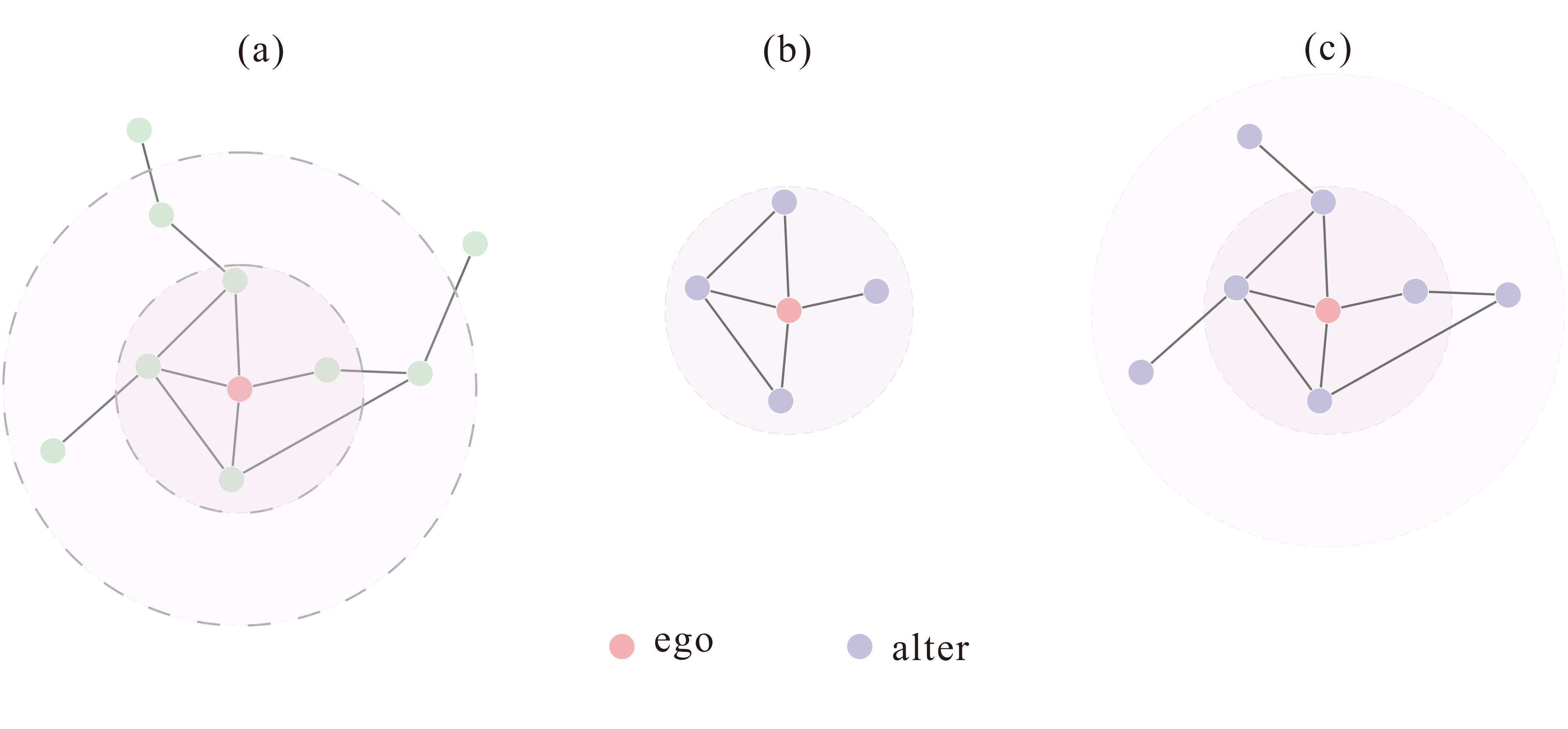}
	\caption{Illustration of the ego network. (a) graph; (b) 1-ego network of the red node; (c) 2-ego network of the red node}
	\label{fig:ego}
\end{figure}

\subsection{Topological Signal Extraction}
After obtaining the ego network of nodes, the structural equivalence between node pairs can be compared based on this network. There are many ways to compare structural equivalence. Here, we calculate the structural equivalence based on the topological signal extracted by persistent homology. The topological features extracted by persistent homology can 1) encode different levels of topological information from the original data, which can describe richer local features than other metrics. Hajij et al.\cite{hajij2021persistent} show that combining the encoding obtained by 
persistent homology with other invariant encoding obtained directly from node embedding can improve the quality of downstream learning tasks; 2) the persistent homology extracts features that are invariant under spatial transformation and more robust to noise.

In order to apply persistent homology on the ego networks, first, the information on the graph needs to be converted into real values, i.e., a function $f: G\to R$ is needed to assign values to nodes and edges. When the function is targeted at nodes, it can be formed as $f: V\to R$, and then the value assigned to an edge can be computed from the nodes it connected, as Equation \ref{eq:node2edge}:
\begin{equation}
  f(e_{ij}) = \max\{f(i),f(j)\}
  \label{eq:node2edge}
\end{equation}

When the function is applied on edges, it can be formed as $f: E\to R$, and the value assigned to a node is based on the edges it linked, as Equation \ref{eq:edge2node}:
\begin{equation}
  f(i) = \min_{i \in e,e\in E}f(e)
  \label{eq:edge2node}
\end{equation}

\begin{figure}[htpb]
	\centering\includegraphics[width=1\linewidth]{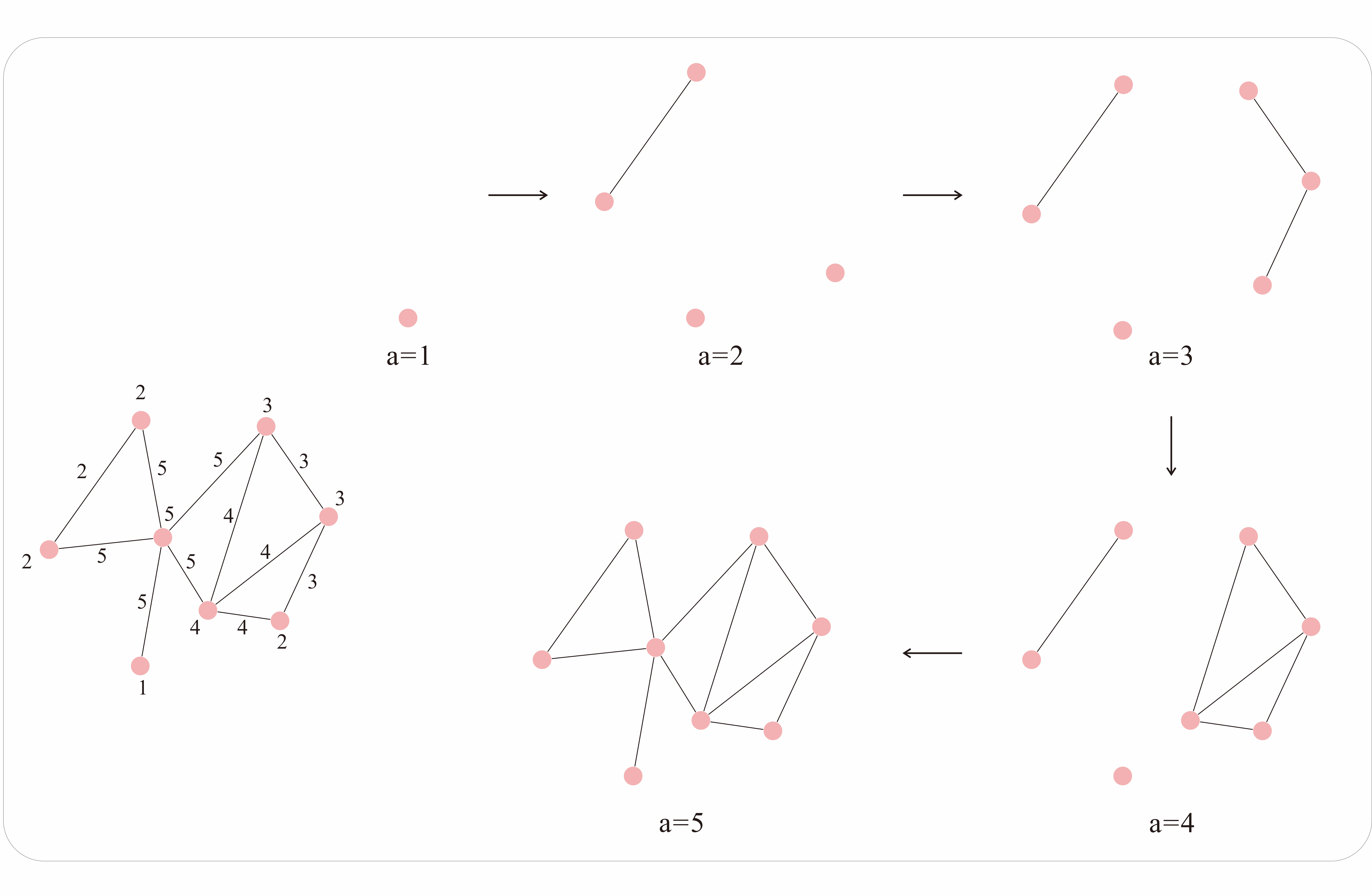}
	\caption{Filtration based on the value of nodes and edges}
	\label{fig:nweighted}
\end{figure}

Then, the simplicial complex at threshold $a$ can be derived from nodes and edges induced from Equation \ref{eq:subset}. This process is illustrated in Figure \ref{fig:nweighted}.
\begin{equation}
G_{\leq a}:=\{\sigma \in V \cup E|f(\sigma)\leq a\}
\label{eq:subset}
\end{equation}

$f$ can be chosen from several options, either important features of nodes on the graph or properties of edges. Here, we target edges, and the function we choose is the Ricci curvature. Among all the definitions of Ricci curvature on the graph, we adopt the widely used one, Ollivier Ricci curvature \cite{ollivier2007ricci,ni2015ricci}. The Ollivier Ricci curvature is formalized as Equation \ref{eq:ricci_c}. It indicates the frequency of triangles and can be used to measure the overlaps of neighbors between nodes, which is highly correlated to the local clustering coefficient \cite{jost2014ollivier}.
\begin{equation}
  \kappa(i,j)=1-\frac{W\left(m_i, m_j\right)}{d(i, j)},
  \label{eq:ricci_c}
  \end{equation}
where $i$ and $j$ denote the nodes, and $d(i,j)$ is the distance on graph. If $i$ and $j$ are adjacents, then $d(i,j)$ is one; otherwise, the distance is determined by the shortest path distance. $m_i$ and $m_j$ are the probability distribution of  $i$ and $j$, it is defined as Equation \ref{eq:ricci2}.
\begin{equation}
  m_i(u)=\left\{\begin{array}{cl}
  \alpha & \text { if } u=i \\
  (1-\alpha) / k & \text { if } u \in N(i) \\
  0 & \text { otherwise }
  \end{array}\right.
  \label{eq:ricci2}
\end{equation}
where $\alpha$ is used to adjust the weight distribution and is normally set to 0.5 \cite{ni2015ricci}.

$W\left(m_i, m_j\right)$ is the optimal transportation distance between $m_i$ and $m_j$, and it is computed as \ref{eq:wa_graph}:
\begin{equation}
  W\left(m_i, m_j\right):=\inf _{\xi \in \prod\left(m_i, m_j\right)} \int \int d(i, j) d \xi(i, j)
  \label{eq:wa_graph}
  \end{equation}
$d \xi(i, j)$ is the sum of mass transported from probability distribution $m_i$ to $m_j$, while $m_i$ and $m_j$ can all be regarded as objects with mass one. $ \prod\left(m_i, m_j\right)$ represents different transport schemes from $m_i$ to $m_j$. $W\left(m_i, m_j\right)$ aims to characterize the overlaps between the local topological structures of two nodes. When $W\left(m_i, m_j\right)$ is smaller than $d(i,j)$, it means the neighborhoods are highly overlapped.

Based on the real values assigned by $f$, we can apply persistent homology on each ego network and extract its topological signal, which then transforms into the persistence image and fed into the representation learning model.

\subsection{Contrastive Learning Based on Structural Equivalence}
After extracting the R-ego subgraph of each node, the topological information of each subgraph can be calculated and converted to PI, and then the structural equivalence between node pairs can be defined by the similarity between PI pairs and the hop distance between node pairs on the graph. Define a positive sample with high structural equivalence as d+, and the nodes u and v in this positive sample pair should satisfy the following conditions.

That is, directly adjacent node pairs are not considered in the calculation of structural equivalence (setting the hop limit $\delta$), and we consider their structural equivalence high only if they are less than a certain distance threshold. To draw the pairs with high structural equivalence close in the feature space, the loss function is designed as follows.
\begin{equation}
    d(u,v)\leq \epsilon \text{  and  } hop(u,v) \geq \delta.
\end{equation}

Finally, based on our assumption (i.e., the introduction of topological signals is an information complement, and homophily and structural equivalence should be expressed as a reasonable combination), joint training is used to incorporate topological information in an adjustable way, which can automatically adjust the importance of topological supervision according to the dataset and the task.
\begin{equation}
    L_{SSL} = -\log \frac{\exp (d_+v^Tu/\tau)}{\exp (d_-v^Tu/\tau)}.
\end{equation}
That is, the final loss can be expressed as $Loss = L_{task} + \lambda L_{SSL}$, where $L_{task}$ is the loss function for a specific downstream task. In other words, our goal is to optimize both the self-supervised loss (i.e., $L_{SSL}$) and the supervised loss (i.e., $L_{task}$).
\begin{figure*}[htpb]
	\centering\includegraphics[width=1\linewidth]{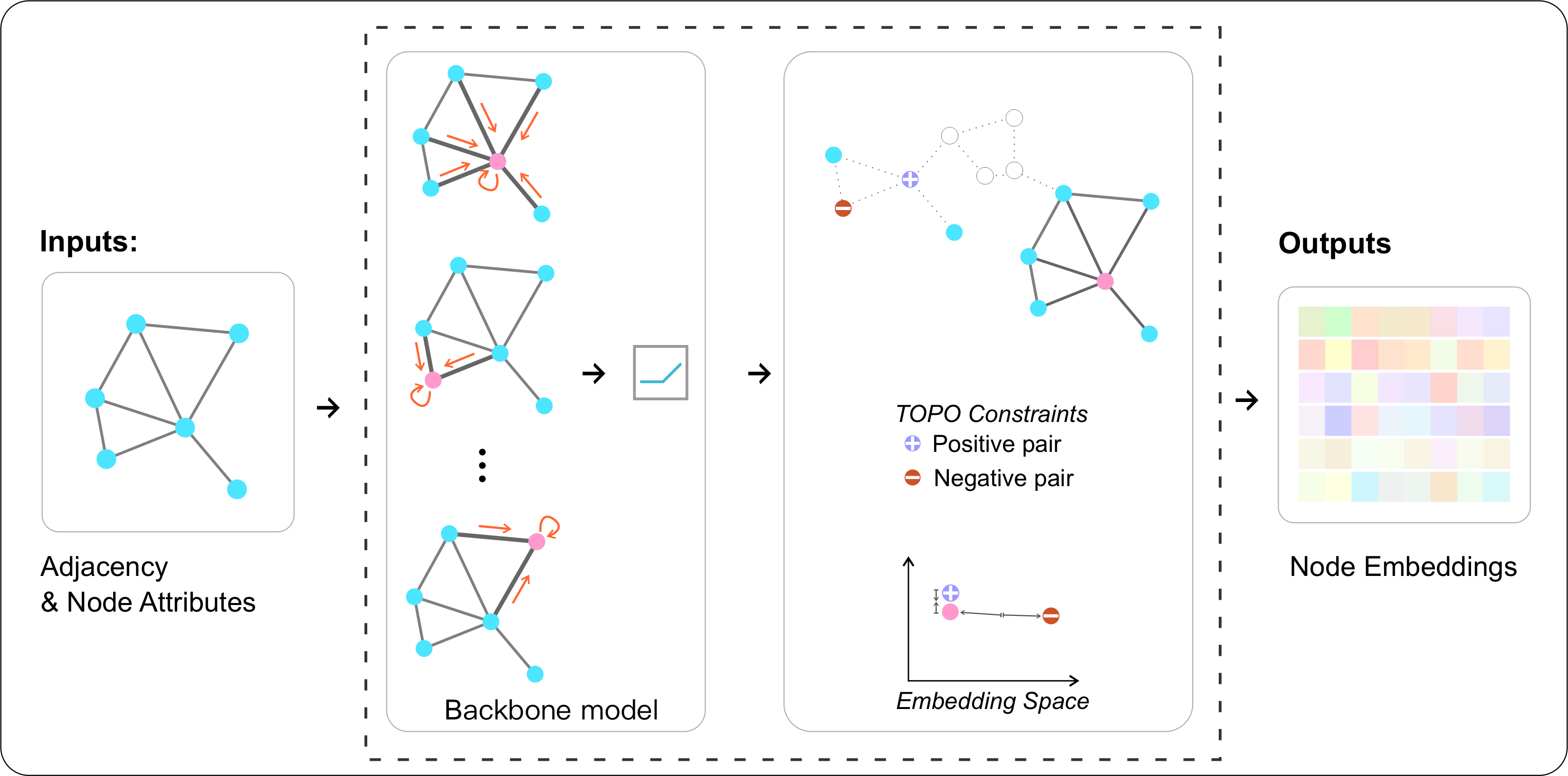}
	\caption{Workflow}
	\label{fig:sslarchitecture}
\end{figure*}

\section{Experiments}
\subsection{Datasets}
We conduct a large number of experiments on seven node classification benchmark datasets, including the citation datasets Cora, Citeseer, PubMed, the paper co-author datasets Coauthor CS, Coauthor Physics, the co-author datasets Amazon Computers, and Amazon Photo. Nodes of Cora, Citeseer, and PubMed datasets are papers, edges indicate the existence of citation relationship, and node categories indicate the field of the paper. Nodes of CS and Physics graphs indicate authors, edges indicate collaboration relationships, and categories indicate the author's research field. Computers and Photo are Amazon co-purchase networks, with nodes represent goods and edges represent connected goods are frequently co-purchased. The statistics of these seven node classification benchmark datasets are shown in the Table \ref{tbl:details}.
\begin{table}[h]
  \centering
  \caption{Dataset details}
  \begin{tabular}{cccc} 
  \hline
  Dataset          & Classes & Nodes & Edges   \\ 
  \hline
  Cora             & 7       & 2485  & 5069    \\
  Citeseer         & 6       & 2110  & 3668    \\
  PubMed           & 3       & 19717 & 44324   \\
  Amazon Computers & 8       & 7487  & 245778  \\
  Amazon Photo     & 10      & 13381 & 119043  \\
  Coauthor Physics & 5       & 34493 & 495924  \\
  Coauthor CS      & 15      & 18333 & 81894   \\
  \hline
  \end{tabular}
  \label{tbl:details}
  \end{table}
\subsection{Experimental Settings}
Our proposed method is a pluggable component, and we choose two models, GCN and GAT, as the backbone models and test the improvement of our proposed method on the backbone model on the node classification task. For both models, we choose a two-layer graph convolution with 1000 training epochs and stop training if the loss function of the model does not change in 200 rounds.
\subsection{Experimental Results}
\begin{table*}[htbp]
  \centering
  \caption{Experimental results}
  \begin{tabular}{ccccccc} 
  \hline
  Dataset       & GAT  & \begin{tabular}[c]{@{}c@{}}TOPOGSSL\\(GAT)\end{tabular} &Improvement     & GCN  & \begin{tabular}[c]{@{}c@{}}TOPOGSSL\\(GCN)\end{tabular} & Improvement      \\ 
  \hline
  Cora      & 82.7 & 84.3                                                    & 1.93\% & 80.6 & 83.7                                                    & 3.85\%  \\
  Citeseer  & 71.6 & 73.3                                                    & 2.37\% & 70.2 & 72.5                                                    & 3.28\%  \\
  PubMed    & 77.8 & 78.6                                                    & 1.03\% & 78.1 & 79                                                      & 1.15\%  \\
  Amazon Computers & 83.1 & 84.5                                                    & 1.68\% & 80.3 & 83.4                                                    & 3.86\%  \\
  Amazon Photo     & 92.1 & 93.7                                                    & 1.74\% & 90.6 & 92.3                                                    & 1.88\%  \\
  Coauthor Physics   & 92.2 & 92.4                                                    & 0.22\% & 92.1 & 92.4                                                    & 0.33\%  \\
  Coauthor CS        & 91.2 & 92.1                                                    & 0.99\% & 90.7 & 91.7                                                    & 1.10\%  \\
  \hline
  \end{tabular}
  \label{tbl:3}
  \end{table*}
  The experimental results are shown in Table \ref{tbl:3}. It can be found that our method can bring improvement regardless of whether GAT or GCN is used as the backbone model, and the greatest improvement can reach 3.85\%. We also observe that although GCN does not perform as well as GAT on some datasets, TOPOGSSL (GCN) outperforms GAT after considering structural equivalence. That is, even when the backbone model performs poorly, using our strategy allows it to achieve competitive results. Also, we observe that the homophily index and the model improvement do not have a linear relationship, but the datasets with low homophily index (e.g., Citeseer and Computer datasets) both improve more, reaching 2.37\%/3.28\% and 1.68\%/3.86\%, respectively. The Physics dataset has the least improvement with only 0.22\%/0.33\%, which can be explained by its high homogeneity index (93.13\%), as less than 7\% of its positive sample pairs are not neighbors.

\section{Discussion}
\subsection{Definition of Distant Samples}
In defining structural equivalence, we consider remote nodes only to be taken into account. We used the parameter $\delta$, i.e., when deriving positive sample pairs based on structural equivalence, we only considered node pairs with distances greater than $\delta$ hops. We analyze the parameter $\delta$, and the results are shown in Table \ref{tbl:4}. It can be found that although the best performing $\delta$ is not fixed for each dataset, the performance of the model improves over the original model on all datasets regardless of which value we choose from [3,4,5].
\begin{table}[htbp]
  \centering
  \caption{The performances under different $\delta$ settings}
  \label{tbl:4}
  \begin{tabular}{cccc} 
  \hline
  Dataset\textbf{} & 3\textbf{} & 4\textbf{} & 5\textbf{}  \\ 
  \hline
  Cora             & 83.7       & 84.1       & \textbf{84.3}        \\
  Citeseer         & 72.8       & 72.6       & \textbf{73.3}        \\
  PubMed           & 78.3       & \textbf{78.6}      & 78.2        \\
  Amazon Computers        & 84.4       & 84.4       & 84.4        \\
  Amazon Photo            & 93.2       & 92.6       & \textbf{93.4}        \\
 Coauthor Physics          & \textbf{92.4}       & \textbf{92.4}       & 92.3        \\
 Coauthor CS               & \textbf{92.1}       & 91.9       & 91.8        \\
  \hline
  \end{tabular}
  \end{table}

\subsection{The Importance of Structural Equivalence}
Since we use a joint training strategy and the parameter $\lambda$ to regulate the effect of structural equivalence on the model, here, we will vary the value of this parameter to explore its effect on the model. The results are shown in Table \ref{tbl:5}, where the bolded ones are the optimal performance on different datasets. We find that a smaller $\lambda$ results in better result. This is because most data sets have a homophily index greater than 0.7. This phenomenon is more evident in the Physics data set (which has the highest homophily index), where the performance decreases and is lower than the original model when the $Loss_{topo}$ ratio is greater than 0.1. Combining the results of this experiment and the homophily index statistics, we believe that a smaller $\lambda$ should be used when using this method on other datasets.
\begin{table}[h]
  \centering
  \caption{The performances under different $\lambda$ settings}
  \begin{tabular}{ccccccc} \hline
  {Dataset} & {0.1}  & {0.3}                                  & {0.5}                                  & {0.7}                                  & {0.9} & {1}  \\ 
  \hline
  Cora         & \textbf{84.3} & 83.7                                          & 83.1                                          & \textit{{81.1}} & 78.7         & 75.6        \\
  Citeseer     & \textbf{73.3} & 72.2                                          & \textit{{71.0}} & 68.4                                          & 63.0         & 61.0        \\
  PubMed       & \textbf{78.2} & 78.0                                          & \textit{{77.1}} & 75.1                                          & 73.6         & 71.7        \\
  Amazon Computers    & \textbf{84.4} & 83.1                                          & \textit{{80.0}} & 78.2                                          & 73.0         & 72.9        \\
  Amazon Photo        & 93.4          & \textbf{93.7}                                 & 92.8                                          & 93.0                                          & 92.7         & 92.7        \\
 Coauthor Physics      & \textbf{92.3} & \textit{{91.9}} & 91.7                                          & 91.7                                          & 90.6         & 91.0        \\
 Coauthor CS           & \textbf{91.8} & 91.8\textbf{}                                 & \textit{{91.1}} & 91.5                                          & 91.0         & 90.8        \\
  \hline
  \end{tabular}
  \label{tbl:5}
  \end{table}

\subsection{The influence of diffrent topological descriptions}
\begin{table}[h]
  \centering
  \caption{The performances under differen $f$}
  \begin{tabular}{cccc} 
  \hline
  Dataset       & $f(ricci) $     & $f(degree)   $  & Original performance  \\ 
  \hline
  Cora      & \textbf{84.3} & 84.2          & 82.7    \\
  Citeseer  & \textbf{73.3} & 72.9          & 71.6    \\
  PubMed    & 78.6          & \textbf{78.8} & 77.8    \\
  Amazon Computers & 84.5          & \textbf{84.8} & 83.1    \\
  Amazon Photo     & \textbf{93.7} & 93            & 92.2    \\
  Coauthor Physics   & 92.4\textbf{} & \textbf{92.5} & 92.1    \\
  Coauthor CS        & 92.1          & 92.1\textbf{} & 91.2    \\
  \hline
  \end{tabular}
  \label{tbl:6}
  \end{table}

(a) In extracting local topological information of nodes using persistent homology. Choosing from different functions is based on the local information under consideration. Here we compare the results of using node degree and Ricci curvature as assignment methods, respectively. As shown in Table \ref{tbl:6}, either method can improve the model significantly. However, the performance improvement does not vary greatly, indicating that the method benefits from the fact that persistent homology can characterize complex local topological information of nodes and thus is not sensitive to the specific assignment method.

\begin{figure}[ht]
	\centering\includegraphics[width=1\linewidth]{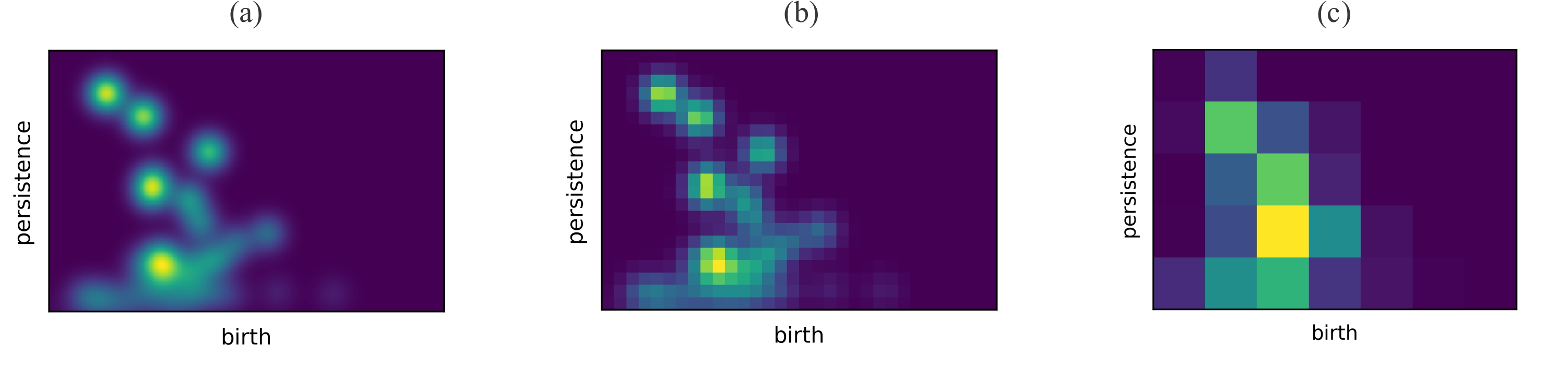}
	\caption{PI generated under different resolution settings}
	\label{fig:pires}
\end{figure}

(b) In vectorizing the results of persistent homology into a persistent image, an important parameter is involved: the resolution. That is, what granularity is used to describe the topology, as shown in Figure \ref{fig:pires}. We use two parameter settings for comparison, and the results are shown in Table \ref{tbl:7}. It can be found that on smaller (small number of edges, nodes) and sparse (low edge/node ratio) datasets, using finer granularity to describe the topology provides better results, while on large and dense datasets, using coarse characterization granularity instead is better. However, the model is improved regardless of whether coarse or fine granularity is used.

\begin{table}[h]
  \centering
  \caption{The performances under different resolutions}
  \label{tbl:7}
  \begin{tabular}{cccc} 
  \hline
  Dataset          & 0.05          & 0.1           & Edges/Nodes  \\ 
  \hline
  Cora             & \textbf{84.3} & \textbf{84.3} & 2.0          \\
  Citeseer         & \textbf{73.3} & 72.6          & 1.7          \\
  PubMed           & \textbf{78.6} & 78.4          & 2.2          \\
  Amazon Computers & 84.5          & \textbf{85.0} & 18.4         \\
  Amazon Photo     & 93.7          & \textbf{93.9} & 15.9         \\
  Coauthor Physics          & 92.4          & \textbf{92.5} & 14.4         \\
  Coauthor CS               & \textbf{92.1} & 91.9          & 4.5          \\
  \hline
  \end{tabular}
  \end{table}

\section{Conclusion}

In recent years, self-supervised graph convolutional neural networks based on the message-passing mechanism have been continuously proposed. This mechanism enables graph representation learning by aggregating messages from neighboring nodes. Its success lies in that most identically labeled nodes tend to produce links in real-world networks, known as homophily. However, our observation of real-world datasets reveals that in addition to the homophily-derived positive samples, there exists a fraction of positive samples that are distant on the graph but have a high topological structural equivalence. A model based on the message-passing mechanism ends up with neighbor bias and is not able to capture the positive samples derived from structural equivalence.

To alleviate neighbor bias, this study proposes a self-supervised method that considers structural equivalence, which uses persistent homology to capture the topological signals inherent in the data. The reason we use persistent homology is its ability to extract topological features at multiple scales. After calculating the structural equivalence of node pairs based on the extracted topological features, we pull in non-direct neighbor node pairs with high structural equivalence in the representation space. The joint training strategy is used to adjust the effect of structural equivalence on the model to fit datasets with different characteristics. We conduct experiments on seven datasets. The results show that the proposed method is effective in improving the model performance, and the model gains from using different perspectives and granularity for local topology characterization. This demonstrates the great potential that the portrayal of the complex topological signals of the graph itself can bring to graph representation learning.

\appendices


\ifCLASSOPTIONcompsoc
  \section*{Acknowledgments}
\else
  \section*{Acknowledgment}
\fi

This work was supported in part by the National Natural Science Foundation of China, Grant/Award Numbers 41871364 and 41871276, the High Performance Computing Platform of Central South University and HPC Central of Department of GIS, in providing HPC resources.

\ifCLASSOPTIONcaptionsoff
  \newpage
\fi



\bibliographystyle{IEEEtran}
\bibliography{ref.bib}
%



%




\end{document}